\documentclass[a4paper,10pt]{article}

\usepackage{amsmath}
\usepackage{amsfonts}
\usepackage{amsthm}
\usepackage{algorithm}
\usepackage[noend]{algpseudocode}
\usepackage{listings}
\usepackage{delarray}
\usepackage{graphicx}
\usepackage{url}
\usepackage{enumerate}
\usepackage{listings}
\usepackage{hyperref}

\newcommand\cmd[1]{\texttt{#1}} % ukaze, ipd. pisi s typewriter stilom  
\newtheorem{definition}{Definition}
\newtheorem{lemma}{Lemma}
\newtheorem{claim}{Claim}
\newtheorem{theorem}{Theorem}

\author{Bla\v{z} Sovdat\footnote{Artificial Intelligence Laboratory, Jo\v{z}ef Stefan Institute, Jamova 39, 1000 Ljubljana, Slovenia. E-mail: \cmd{blaz.sovdat@gmail.si}.}}
\title{Updating Formulas and Algorithms for Computing Entropy and Gini Index from Time-Changing Data Streams}

\begin{document}
	\maketitle

	\begin{abstract}
		Despite growing interest in data stream mining the most successful incremental learners still use periodic recomputation to update attribute information gains and Gini indices. This note provides simple incremental formulas and algorithms for computing entropy and Gini index from time-changing data streams.
	\end{abstract}

	\section{Introduction}
		Information-theoretic entropy was introduced by Shannon in the celebrated 1948 paper~\cite{shannon48} and has since found a vast number of applications~\cite{elements2006}. In machine learning, information gain~\cite{mitchell1997ml}, defined as the expected entropy reduction after splitting a leaf on a given attribute, is one of the most popular impurity measures for decision tree learning.

		However, within the data stream mining world, there is a need for computationally cheap update formulas --- an alternative being the complete and expensive recomputation --- to compute the entropy as new examples come in, changing the sample distribution. Concrete example of such scenario is found in the incremental decision tree learners VFDT~\cite{domingos2000mining} and CVFDT~\cite{hulten2001mining}. The same holds for Gini index, another popular impurity measure used for decision tree and IF-THEN rule learning. 

		The main contributions are twofold.
		\begin{itemize}
			\item Theorems~\ref{thm:gini_inc} and~\ref{thm:gini_conc} that give simple update formulas for Gini index as new examples enter the stream (e.g. of unseen type) in and as some of the sample counts change, respectively, and algorithms that estimate ``current'' Gini index of the data stream using our formulas with sliding windows (Algorithm~\ref{algo:gini_w}) and fading factors (Algorithm~\ref{algo:gini_ff}).
			\item Theorems~\ref{thm:en} and~\ref{thm:ch} that give simple update formulas for entropy as new examples enter the stream and as some of the sample counts change, and  algorithms that estimate the current entropy of the data stream with sliding windows (Algorithm~\ref{algo:entropy_w}) and fading factors (Algorithm~\ref{algo:entropy_ff}).
		\end{itemize}
		Despite their simplicity and growing importance of the data stream model, we are not aware of update formulas for information gain and Gini index.

		% The main contribution are theorems~\ref{thm:gini_inc} and~\ref{thm:gini_conc} that give simple update formulas for Gini index as new examples come in and as some of the sample counts change, respectively, and algorithms~\ref{algo:gini_w} and~\ref{algo:gini_ff} that estimate ``current'' Gini index of the data stream using our formulas with sliding windows and fading factors. We do the same for entropy in theorems~\ref{thm:en} and~\ref{thm:ch} and algorithms~\ref{algo:entropy_w} and~\ref{algo:entropy_ff}. Despite their simplicity and growing importance of the data stream model, we are not aware of update formulas for information gain and Gini index.

		The results are not general as we assume that we are incrementally computing the entropy of distribution of labels in a data stream, where the probabilities can be expressed as fractions of label counts (this is evident from formulas and algorithms which rely on these structural assumptions). However, these assumptions are satisfied in machine-learning scenarios (for example, decision tree learning) that motivated this note.

		For a more theoretical perspective, the interested reader should consult the work of Chakrabarti et al.~\cite{chakrabarti2007near,chakrabarti2006estimating}.

		The note is organized as follows. In Section~\ref{sec:gini} we derive incremental formulas for Gini index and then use these formulas (Subsection~\ref{subsec:gini_algo}) with sliding windows and fading factors to get algorithms for computing Gini index of time-changing data streams. In Section~\ref{sec:entropy} we give analogous formulas for entropy and use them (Subsection~\ref{subsec:entropy_algo}) with sliding windows and fading factors to get algorithms for estimating entropy of time-changing data streams. We conclude the note in Section~\ref{sec:conclude}. % In section~\ref{sec:uses} we outline several use cases for the derived formulas and algorithms. We conclude the note in section~\ref{sec:conclude}.

	\section{Incremental Formulas for Gini Index}\label{sec:gini}
		Let $\{x_i\}_{i=1}^n$ be a sample of positive real numbers and let $S_n:=x_1+x_2+\ldots+x_n$ be sum of sample elements. Letting $p_i:=x_i/S_n$ we get a discrete distribution. In what follows we work with samples of positive real numbers (modeling a data stream) and are interested in computing their Gini index (defined below) or entropy (defined in the next section), that is, computing the Gini index or the entropy of the distribution formed by $p_i$'s.
		\begin{definition}
			Gini index of sample of real numbers $\{x_i\}_{i=1}^n$ is defined as 
			\begin{equation*}
				G_n:=1-\sum_{i=1}^n p_i^2 = 1-\sum_{i=1}^n\left(\frac{x_i}{S_n}\right)^2=1-\frac{1}{S_n^2}\sum_{i=1}^n x_i^2.
			\end{equation*}
		\end{definition}
		
		We will use the following obvious equality throughout this section.
		\begin{lemma}\label{lemma:gini}
			Let $\{x_i\}_{i=1}^n$ be a sample of positive real numbers, let $S_n$ be the sum of sample element, and let $G_n$ be the sample Gini index. We then have
			\begin{equation}
				\sum_{i=1}^n x_i^2 = S_n^2(1-G_n).
			\end{equation}
		\end{lemma}
		\begin{proof}
			Write out the right-hand side and do the algebra:
			\begin{align*}
				S_n^2(1-G_n) &= S_n^2\left(1+\frac{1}{S_n^2}\sum_{i=1}^n x_i^2-1\right) = \sum_{i=1}^n x_i^2.
			\end{align*}
		\end{proof}
		
		The next claim gives update formula when one of the sample elements increases by one.
		\begin{claim}\label{claim:gini_inc}
			Let $\{x_i\}_{i=1}^n$ be sample of positive real numbers and let $G_n$ be sample Gini index. Suppose $x_i$ changes to $x_i+1$ and let $G_n'$ denote Gini index of the new sample. We then have
			\begin{equation}\label{eq:gini_one}
				G_n'=1-\frac{1}{(S_n+1)^2}\left(S_n^2\left(1-G_n\right)+2x_i+1\right).
			\end{equation}
		\end{claim}
		\begin{proof}
			Plug the new value for $x_i$ and do the algebra:
			\begin{align*}
				G_n' &= 1-\frac{1}{S_n'^2}\sum_{i=1}^n x_i'^2 \\
					&= 1-\frac{1}{(S_n+1)^2}\left(\sum_{i=1}^n x_i^2 + 2x_i+1\right) \\
					&= 1-\frac{1}{(S_n+1)^2}\left(S_n^2\left(1-G_n\right)+2x_i+1\right).
			\end{align*}
		\end{proof}
		
		The next theorem generalizes Claim~\ref{claim:gini_inc}.
		\begin{theorem}\label{thm:gini_inc}
			Let $\{x_i\}_{i=1}^n$ be sample of positive real numbers and let $G_n$ be sample Gini index. Suppose that $i$th sample element $x_i$ increases by $r_i>0$ for $i\in I$, where $I$ is index set (indices of elements that change). Define $r:=r_1+r_2+\ldots+r_n$ with $r_i:=0$ for $i\notin I$. We then have
			\begin{equation}
				G_n'=1-\frac{1}{(S_n+r)^2}\left(S_n^2\left(1-G_n\right)+\sum_{i\in I}\left(2x_ir_i+r_i^2\right)\right).\label{eq:gini_inc}
			\end{equation}
		\end{theorem}
		\begin{proof}
			Similarly as before, we do the algebra to get the result:
			\begin{align*}
				G_n' &= 1-\frac{1}{(S_n+r)^2}\sum_{i=1}^n (x_i+r_i)^2 \\
						&= 1-\frac{1}{(S_n+r)^2}\sum_{i=1}^n\left(x_i^2+2x_ir_i+r_i^2\right) \\
						&= 1-\frac{1}{(S_n+r)^2}\left(S_n^2\left(1-G_n\right)+\sum_{i\in I}\left(2x_ir_i+r_i^2\right)\right),
			\end{align*}
			where the last equality follows by Lemma~\ref{lemma:gini}.
		\end{proof}
		Note that the number of required operations in Equation~\eqref{eq:gini_inc} grows linearly with the number of elements that changed, i.e., it takes $O(|I|)$ operations to update the Gini index. 
		
		The next claim gives update formula for Gini index when a new element enters the sample.
		\begin{claim}\label{claim:gini_add}
			Let $\{x_i\}_{i=1}^n$ be sample of positive real numbers and let $S_n$ be the sum of sample elements and let $G_n$ be the sample Gini index. Suppose new element $x_{n+1}$ enters the sample. Gini index then becomes
			\begin{equation}
				G_{n+1}=1-\frac{1}{(S_n+x_{n+1})^2}\left(S_n^2(1-G_n)+x_{n+1}^2\right).
			\end{equation}
		\end{claim}
		\begin{proof}
			We clearly have
			\begin{align*}
				G_{n+1} &= 1-\frac{1}{(S_n+x_{n+1})^2}\left(\sum_{i=1}^nx_i^2+x_{n+1}^2\right) \\
					&= 1-\frac{1}{(S_n+x_{n+1})^2}\left(S_n^2(1-G_n)+x_{n+1}^2\right),
			\end{align*}
			where the last equality follows by Lemma~\ref{lemma:gini}.
		\end{proof}
		
		The next theorem generalizes Claim~\ref{claim:gini_add}, giving update formula for Gini index when we ``concatenate'' two samples.
		\begin{theorem}\label{thm:gini_conc}
			Let $\{x_i\}_{i=1}^m$ and $\{y_i\}_{i=1}^n$ be samples of positive real numbers and let $R_m$ and $S_n$ be sums of sample elements. Furthermore, let $F_m$ and $G_n$ be sample Gini indices and define
			\begin{align*}
				z_i & :=
				\left\{\begin{array}{cl}
					x_i, & 1\le i\le m, \\
					y_{i-m}, & m+1\le i\le m+n.
				\end{array}\right.
			\end{align*}
			Then the Gini index becomes
			\begin{equation*}
				E_{m+n}=1-\frac{1}{(R_m+S_n)^2}\left(R_m^2(1-F_m)+S_n^2(1-G_n)\right).
			\end{equation*}
		\end{theorem}
		\begin{proof}
			By definition we have
			\begin{align*}
				E_{n+m}&=1-\frac{1}{(R_m+S_m)^2}\sum_{i=1}^{n+m}z_i^2 \\
					&= 1-\frac{1}{(R_m+S_n)^2}\left(\sum_{i=1}^m x_i^2+\sum_{i=1}^n y_i^2\right) \\
					&= 1-\frac{1}{(R_m+S_n)^2}\left(R_m^2(1-F_m)+S_n^2(1-G_n)\right),
			\end{align*}
			where the last equality follows by applying Lemma~\ref{lemma:gini} twice.
		\end{proof}
		
		\begin{theorem}\label{thm:gini_cup}
			Let $\{x_i\}_{i=1}^n$ and $\{y_i\}_{i=1}^n$ be samples of positive real numbers, let $R_n$ and $S_n$ be sums of sample elements, and let $F_n$ and $G_n$ be sample Gini indices. Furthermore, let $z_i:=x_i+y_i$ for $1\le i\le n$. Then Gini index of the ``union'' becomes
			\begin{equation*}
				E = 1-\frac{1}{(R_n+S_n)^2}\left(R_n^2\left(1-F_n\right)+S_n^2\left(1-G_n\right)+2\sum_{i=1}^n x_iy_i\right)
			\end{equation*}
		\end{theorem}
		\begin{proof}
			By definition we have
			\begin{equation*}
				F_n = 1-\frac{1}{R_n^2}\sum_{i=1}^n x_i^2
			\end{equation*}
			and
			\begin{equation*}
				G_n = 1-\frac{1}{S_n^2}\sum_{i=1}^n y_i^2.
			\end{equation*}
			First part of the formula follows by Theorem~\ref{thm:gini_inc}, while the last part, $2(x_1y_1+x_2y_2+\ldots+x_ny_n)$, is not ``stored'' anywhere and has to be recomputed.
		\end{proof}
		
		\subsection{Algorithms for Computing Gini Index on Time-Changing Data Streams}\label{subsec:gini_algo}
			Data streams are inherently changing and we are usually interested in ``recent'' Gini index. In this section we propose two algorithms for this problem --- Algorithm~\ref{algo:gini_w} uses sliding windows, while Algorithm~\ref{algo:gini_ff} uses fading factors to capture the ``recent'' Gini index. 
			
			Algorithm~\ref{algo:gini_w} computes Gini index of the last $w\in\mathbb{N}$ stream elements. It achieves this using sliding window of size $w$, meaning its space complexity is $O(w)$. Note that $w$ is user-defined parameter, which indicates what subset of stream elements is ``recent''.
			\begin{algorithm}
				\caption{Computing Gini index using sliding windows.}
				\label{algo:gini_w}
				\begin{algorithmic}[1]
					\Require Sliding window size $w\in\mathbb{N}$ and a data stream $S$.
					\Ensure Be ready to return the Gini index of the sliding window at any time.
					\State Let $W:=\{\}$ be a sliding window.
					\State Let $n:=0$ be the number of all examples.
					\State Let $n_i:=0$ be the number of examples from the $i$-th class.
					\State Let $g:=0$ be the current Gini index. 
					\For{$x\in S$} % \Comment{TODO: Add usage example for $\Call{Add}{}$ and $\Call{Del}{}$}
						\If{$|W|>w$}
							\State Remove the oldest element $x'$, labeled with the $i$-th class, from the sliding window $W$.
							\State Update $g := \Call{Dec}{g, n, n_i}$.
						\EndIf
						\State Add $W:=W\cup\{x\}$ element labeled with the $i$-th class.
						\State Update $g := \Call{Inc}{g, n, n_i}$.
					\EndFor
					\Function{Add}{$g$, $n$, $n_i$} \Comment{Append}
						\State Update $n:=n+n_i$
						\State \Return $\displaystyle 1-\frac{1}{n^2}\left((n-n_i)^2(1-g)+n_i^2\right)$
					\EndFunction
					\Function{Del}{$g$, $n$, $n_i$} \Comment{Delete}
						\State Update $n:=n-n_i$
						\State \Return $\displaystyle 1-\frac{1}{n^2}\left((n+n_i)^2(1-g)-n_i^2\right)$
					\EndFunction
					\Function{Inc}{$g$, $n$, $n_i$} \Comment{Increment}
						\State Update $n := n+1$ in $n_i := n_i+1$
						\State \Return $\displaystyle 1-\frac{1}{n^2}\left((n-1)^2(1-g)+2n_i-1\right)$
					\EndFunction
					\Function{Dec}{$g$, $n$, $n_i$} \Comment{Decrement}
						\State Update $n:=n-1$ in $n_i:=n_i-1$
						\State \Return $\displaystyle 1-\frac{1}{n^2}\left((n+1)^2(1-g)-2n_i-1\right)$
					\EndFunction
				\end{algorithmic}
			\end{algorithm}
			
			Algorithm~\ref{algo:gini_ff} computes ``recent'' Gini index using fading factors --- element contributions are weighted with $\{\alpha^k:k\in\mathbb{N}\}$ for some fixed $\alpha\in(0,1]$ according to element's ``age''. Note that the fading factor $\alpha$ defines what recent means and that this algorithm has small constant space complexity.
			\begin{algorithm}
				\caption{Computing Gini index using fading factors.}
				\label{algo:gini_ff}
				\begin{algorithmic}[1]
					\Require Fading factor $\alpha\in(0, 1]$ and a data stream $S$.
					\Ensure Be ready to return the current Gini index at any time.
					\State Let $n:=0$ be the number of all examples.
					\State Let $n_i:=0$ be the number of examples from the $i$-th class
					\State Let $g:=0$ be the current Gini index.
					\For{$x\in S$}
						\State Suppose $x$ is from the $i$-th class.
						\State Update Gini index $\displaystyle g:= 1-\frac{1}{(n+1)^2}\left(n^2(1-\alpha g)+2n_i+1\right)$.
						\State Update counts $n:=n+1$ and $n_i:=n_i+1$.
					\EndFor
				\end{algorithmic}
			\end{algorithm}
			
			Of course, a number of generalizations are possible, for example a combination of sliding windows and fading factors. 
	
	\section{Incremental Formulas for Entropy}\label{sec:entropy}
		In this section we derive analogous incremental formulas and algorithms for entropy.
		
		We define information entropy as typically used by machine learning practitioners. Recall that the entropy of a sample of positive real numbers $\{x_i\}_{i=1}^n$ is defined as the entropy of the distribution formed by $p_i$'s for $p_i=x_i/S_n$, where $S_n:=x_1+x_2+\ldots+x_n$.
		\begin{definition}
			Let $\{x_i\}_{i=1}^n$ be sample of positive real numbers and let $S_n:=x_1+x_2+\ldots+x_n$ be sum of sample elements. Define the entropy of the sample as
			\begin{equation*}
				H_n:=-\sum_{i=1}^n\frac{x_i}{S_n}\log_2\frac{x_i}{S_n}.
			\end{equation*}
		\end{definition}
		
		We first prove the following technical lemma, which we use throughout this section. 
		\begin{lemma}\label{lema:entrop}
			Let $\{x_i\}_{i=1}^n$ be sample of positive real numbers and let $H_n$ and $S_n$ be sample entropy and sum of sample elements, respectively. For any positive real number $R>0$ we have
			\begin{equation}
				-\sum_{i=1}^n\frac{x_i}{S_n+R}\log_2\frac{x_i}{S_n+R}=\frac{S_n}{S_n+R}\left(H_n-\log_2\frac{S_n}{S_n+R}\right).
			\end{equation}
		\end{lemma}
		\begin{proof}
			Write $\displaystyle\frac{x_i}{R+S_n}=1\cdot\frac{x_i}{R+S_n}=\frac{S_n}{S_n}\frac{x_i}{R+S_n}=\frac{x_i}{S_n}\frac{S_n}{R+S_n}$. We then clearly have
			\begin{align*}
				-\sum_{i=1}^n\frac{x_i}{S_n+R}\log_2\frac{x_i}{S_n+R} &= -\sum_{i=1}^n\frac{x_i}{S_n}\frac{S_n}{S_n+R}\log_2\left(\frac{x_i}{S_n}\frac{S_n}{S_n+R}\right) \\
					&= -\sum_{i=1}^n\frac{x_i}{S_n}\frac{S_n}{S_n+R}\left(\log_2\frac{x_i}{S_n}+\log_2\frac{S_n}{S_n+R}\right) \\
					&= -\frac{S_n}{S_n+R}\left(\sum_{i=1}^n\frac{x_i}{S_n}\log_2\frac{x_i}{S_n}+\sum_{i=1}^n\frac{x_i}{S_n}\log_2\frac{S_n}{S_n+R}\right) \\
					&= \frac{S_n}{S_n+R}\left(H_n-\log_2\frac{S_n}{S_n+R}\sum_{i=1}^n\frac{x_i}{S_n}\right) \\
					&= \frac{S_n}{S_n+R}\left(H_n-\log_2\frac{S_n}{S_n+R}\right). \\
			\end{align*}
		\end{proof}
		
		The next claim gives simple update formula when a new positive real number $x_i>0$ enters the sample.
		\begin{claim}[\cite{MO133986}]\label{claim:entropy}
			Let $H_n$ and $S_n$ be sample entropy and sum of sample elements and suppose that a new positive real number $x_{n+1}>0$ enters the sample. We then have
			\begin{equation}\label{eq:hinc}
				H_{n+1}=\frac{S_n}{S_{n+1}}\left(H_n-\log_2\frac{S_n}{S_{n+1}}\right)-\frac{x_{n+1}}{S_{n+1}}\log_2\frac{x_{n+1}}{S_{n+1}}.
			\end{equation}
		\end{claim}
		\begin{proof}
			By definition we have
			\begin{align*}
				H_{n+1} &= -\sum_{i=1}^{n+1}\frac{x_i}{S_{n+1}}\log_2\frac{x_i}{S_{n+1}} \\
					&= -\frac{x_{n+1}}{S_{n+1}}\log_2\frac{x_{n+1}}{S_{n+1}}-\sum_{i=1}^n\frac{x_i}{S_{n+1}}\log_2\frac{x_i}{S_{n+1}} \\
					&= \frac{S_n}{S_{n+1}}\left(H_n-\log_2\frac{S_n}{S_{n+1}}\right)-\frac{x_{n+1}}{S_{n+1}}\log_2\frac{x_{n+1}}{S_{n+1}},
			\end{align*}
			with the last equality following from Lemma~\ref{lema:entrop}.
		\end{proof}
		
		The next theorem generalizes the claim and gives formula for entropy of ``concatenated'' samples, given sample entropies $G_m$ and $H_n$ and sums of sample elements $R_m$ and $S_n$.
		\begin{theorem}\label{thm:en}
			Let $\{x_i\}_{i=1}^m$ and $\{y_i\}_{i=1}^n$ be samples of positive real numbers and let $R_m:=x_1+x_2+\ldots+x_m$ and $S_n:=y_1+y_2+\ldots+y_n$ be sums of sample elements. Furthermore let $G_m$ and $H_n$ be sample entropies. Define
			\begin{align*}
				z_i & :=
				\left\{\begin{array}{cl}
					x_i, & 1\le i\le m, \\
					y_{i-m}, & m+1\le i\le m+n,
				\end{array}\right.
			\end{align*}
			and let $Z_{m+n}:=z_1+z_2+\ldots+z_{n+m}=R_m+S_n$. We then have
			\begin{equation}\label{eq:update}
				E_{m+n} = \frac{R_m}{Z_{m+n}}\left(G_m-\log_2\frac{R_m}{Z_{m+n}}\right)+\frac{S_n}{Z_{m+n}}\left(H_n-\log_2\frac{S_n}{Z_{m+n}}\right).
				% H_{n+m} = \frac{S_n}{Z_{n+m}}H_n-\frac{S_n}{Z_{n+m}}\log_2\frac{S_n}{Z_{n+m}}+\frac{R_m}{Z_{m+n}}H_m-\frac{R_m}{Z_{m+n}}\log_2\frac{R_m}{Z_{m+n}}.
			\end{equation}
		\end{theorem}
		\begin{proof}
			Similarly as before, we have
			\begin{align*}
				E_{n+m} &= -\sum_{i=1}^{m+n}\frac{z_i}{Z_{m+n}}\log_2\frac{z_i}{Z_{m+n}} \\
					&= -\sum_{i=1}^m\frac{x_i}{Z_{n+m}}\log_2\frac{x_i}{Z_{m+n}}-\sum_{i=1}^n\frac{y_i}{Z_{m+n}}\log_2\frac{y_i}{Z_{m+n}} \\
					&= \frac{R_m}{Z_{m+n}}\left(G_m-\log_2\frac{R_m}{Z_{m+n}}\right)+\frac{S_n}{Z_{m+n}}\left(H_n-\log_2\frac{S_n}{Z_{m+n}}\right),
			\end{align*}
			where the last equality follows by applying Lemma~\ref{lema:entrop} twice. 
		\end{proof}
		Note that Claim~\ref{claim:entropy} is a corollary of Theorem~\ref{thm:en}, if apply Equation~\eqref{eq:update} to $H_n$ and $x_{n+1}$ and think of $x_{n+1}$ as a sample with a single element. 
		
		Theorem~\ref{thm:ch} gives formula for entropy when some of the elements $x_i$ for $i\in I$ increase by $r_i>0$, where $I$ is index set and we let $r_i:=0$ for $i\notin I$. 
		\begin{theorem}\label{thm:ch}
			Let $\{x_i\}_{i=1}^n$ be a sample of positive real numbers and let $S_n$ be sum of sample elements. Let $H_n$ be sample entropy. Suppose $x_i$ increases by $r_i>0$ for $i\in I$ and let $r:=r_1+r_2+\ldots+r_n$ with $r_i:=0$ for $i\not\in I$. Then the entropy $H_n$ becomes
			\begin{equation}\label{eq:ch}
				\frac{S_n}{S_n+r}\left(H_n-\log_2\frac{S_n}{S_n+r}\right)-\sum_{i\in I} \left(\frac{x_i+r_i}{S_n+r}\log_2\frac{x_i+r_i}{S_n+r}-\frac{x_i}{S_n+r}\log_2\frac{x_i}{S_n+r}\right).
			\end{equation}
		\end{theorem}
		\begin{proof}
			The idea is to think of the terms $\frac{x_i+r_i}{S_n+r}$ as new elements, apply Theorem~\ref{thm:en}, and subtract ``old'' elements $\frac{x_i}{S_n+r}$ for $i\in I$. Note that the elements we are subtracting have $S_n+r$ in the denominator---this is because we are subtracting from the updated entropy (i.e. after we have applied Theorem~\ref{thm:en}).
		\end{proof}
		Note that the number of required operations grows linearly with the number of changed elements --- if $|I|=k$ elements change, we only need $O(k)$ operations. Also note that formulas become (numerically) problematic when $x_n$ is small compared to $S_n$.
		
		\subsection{Algorithms for Computing Entropy on Time-Changing Data Streams}\label{subsec:entropy_algo}
			We now give algorithms for computing ``recent'' entropy --- Algorithm~\ref{algo:entropy_w} uses sliding windows, while Algorithm~\ref{algo:entropy_ff} uses fading factors.
			
			Algorithm~\ref{algo:entropy_w}, similarly as its Gini-index-analog, Algorithm~\ref{algo:gini_w}, sliding window size $w\in\mathbb{N}$, which defines what subset of stream elements is recent. The space complexity is clearly $O(w)$.
			
			Ideally we would want an algorithm that adapts sliding window size --- similarly as ADWIN~\cite{bifet2007learning} does --- because $w$ changes with time due to time-changing nature of data streams.
			\begin{algorithm}[h]
				\caption{Computing entropy using sliding windows.}
				\label{algo:entropy_w}
				\begin{algorithmic}[1]
					\Require Sliding window size $w\in\mathbb{N}$ and a data stream $S$.
					\Ensure Be ready to return the entropy of the sliding window at any time.
					
					\State Let $W := \{\}$ be a sliding window and let $h := 0$ be the current entropy.
					\State Let $n := 0$ be the number of all examples.
					\State Let $n_i := 0$ be the number of examples with the $i$-th class label.
					\For{$x\in S$} % \Comment{Missing usage example for $\Call{Add}{}$ and $\Call{Dec}{}$.}
						\If{$|W|>w$}
							\State Remove the oldest element $x'$, from the $j$-th class, from the sliding window $W$.
							\State Update $h := \Call{Dec}{h, n, n_j}$.
						\EndIf
						\State Suppose $x$ is from the $i-$th class.
						\State Add $W:=W\cup\{x\}$.
						\State Update $h:=\Call{Inc}{h, n, n_i}$.
					\EndFor
					\Function{Add}{$h$, $n$, $n_i$}
						\State Update $n:=n+n_i$
						\State \Return $\displaystyle \frac{n-n_i}{n}\left(h-\log_2\frac{n-n_i}{n}\right)-\frac{n_i}{n}\log_2\frac{n_i}{n}$
					\EndFunction
					\Function{Del}{$h$, $n$, $n_i$}
						\State Update $n:=n-n_i$
						\State \Return $\displaystyle \frac{n+n_i}{n}\left(h+\frac{n_i}{n+n_i}\log_2\frac{n_i}{n+n_i}\right)+\log_2\frac{n}{n+n_i}$
					\EndFunction
					\Function{Inc}{$h$, $n$, $n_i$}
						\State Update $n:=n+1$ and $n_i:=n_i+1$
						\State \Return $\displaystyle \frac{n-1}{n}\left(h-\log_2\frac{n-1}{n}\right)-\frac{n_i}{n}\log_2\frac{n_i}{n}+\frac{n_i-1}{n}\log_2\frac{n_i-1}{n}$
					\EndFunction
					\Function{Dec}{$h$, $n$, $n_i$}
						\State Update $n:=n-1$ and $n_i:=n_i-1$
						\State \Return $\displaystyle \frac{n+1}{n}\left(h+\frac{n_i+1}{n+1}\log_2\frac{n_i+1}{n+1}-\frac{n_i}{n+1}\log_2\frac{n_i}{n+1}\right)+\log_2\frac{n}{n+1}$
					\EndFunction
				\end{algorithmic}
			\end{algorithm}
			
			Algorithm~\ref{algo:entropy_ff}, similarly as Algorithm~\ref{algo:gini_ff}, defines ``recent'' using fading factors $\alpha\in(0,1]$. Element contributions are weighted with $\{\alpha^k:k\in\mathbb{N}\}$ according to element ``age''. The algorithm has small constant space complexity. 
			\begin{algorithm}
				\caption{Computing the entropy from time-changing data streams.}
				\label{algo:entropy_ff}
				\begin{algorithmic}[1]
					\Require Fading factor $\alpha\in(0, 1]$ and data stream $S$.
					\Ensure Be ready to return the current entropy at any time.
					
					\State Let $n:=0$ be the number of all examples.
					\State Let $n_i:=0$ be the number of example in the $i$-th class.
					\State Let $h:=0$ be the current entropy.
					\For{$x\in S$}
						\State Suppose the element $x$ has the $i$-th class label.
						\State Update entropy $\displaystyle h:=\frac{n}{n+1}\left(\alpha h-\log_2\frac{n}{n+1}\right)-\frac{n_i+1}{n+1}\log_2\frac{n_i+1}{n+1}+\frac{n_i}{n+1}\log_2\frac{n_i}{n+1}$.
						\State Update counts $n:=n+1$ and $n_i:=n_i+1$
					\EndFor
				\end{algorithmic}
			\end{algorithm}
		
		%\section{Use cases}\label{sec:uses}
		%	This sections outlines several use-cases for derived formulas and algorithms. [\ldots].
		%	\paragraph{Dynamic attribute values.} Assuming data structure describing the attribute can ``extend'' on-the-fly to incorporate new attribute value, then our formulas correctly incorporate the new attribute value when computing information gain or Gini index.
		%	\paragraph{Avoiding recomputation.} VFDT and CVFDT learners (and their derivatives) periodically recompute information gains and Gini indices. Using our formulas this is no longer necessary.
	\section{Conclusion and Future Work}\label{sec:conclude}
		We derived simple incremental formulas and algorithms for computing entropy and Gini index on time-changing data streams. The derivations are elementary and easy to implement.

		Below, we outline several directions for future work.
		\begin{itemize}
			\item Further investigate numerical stability issues of the derived formulas and algorithms for entropy.
			\item Generalize the algorithms based on sliding windows to adapt the window size automatically, for example as \textsc{ADWIN}~\cite{bifet2007learning} does.
			\item Describe several uses cases for our formulas and algorithms, for example: avoiding recomputation in learners such as VFDT and CVFDT, apply the formulas to let decision-tree learners incorporate new attribute values that need not be prespecified, etc.
			\item We also plan to implement an open-source library that will allow one to use the algorithms we derived.
		\end{itemize}

	\section*{Acknowledgments}
		I thank Jean Paul Barddal of the Pont\'{i}ficia Universidade Cat\'{o}lica do Paranan\'{a} for pointing out errors in the updating functions of Algorithms~\ref{algo:entropy_w} and~\ref{algo:entropy_ff} and errors in Theorems~\ref{thm:en} and~\ref{thm:ch}. These have now been corrected.

		The results were derived in the summer of 2013 while I was working on my bachelor's thesis~\cite{sovdat2013algorithms} at the Artificial Intelligence Laboratory at the Jo\v{z}ef Stefan Institute.

		I also thank Marko Robnik-\v{S}ikonja, Martin Vuk, and Zoran Bosni\'{c} of the Faculty of Computer and Information Science of the University of Ljubljana for useful comments and suggestions and to Bla\v{z} Fortuna, Andrej Muhi\v{c}, Jan Rupnik, and Marko Grobelnik from the Artificial Intelligence Laboratory at the Jo\v{z}ef Stefan Institute. Special thanks to Andrej Muhi\v{c} for pointing out issues with numerical stability.
\bibliographystyle{alpha}
\bibliography{refs}

\begin{thebibliography}{CDBM06}

\bibitem[BG07]{bifet2007learning}
Albert Bifet and Ricard Gavalda.
\newblock Learning from {T}ime-{C}hanging {D}ata with {A}daptive {W}indowing.
\newblock In {\em SDM}, volume~7, page 2007. SIAM, 2007.

\bibitem[CCM07]{chakrabarti2007near}
Amit Chakrabarti, Graham Cormode, and Andrew McGregor.
\newblock A {N}ear-{O}ptimal {A}lgorithm for {C}omputing the {E}ntropy of a
  {S}stream.
\newblock In {\em Proceedings of the {E}ighteenth {A}nnual ACM-SIAM {S}ymposium
  on {D}iscrete {A}lgorithms}, pages 328--335. SIAM, 2007.

\bibitem[CDBM06]{chakrabarti2006estimating}
Amit Chakrabarti, Khanh Do~Ba, and S~Muthukrishnan.
\newblock Estimating {E}ntropy and {E}ntropy {N}orm on {D}ata {S}treams.
\newblock {\em Internet {M}athematics}, 3(1):63--78, 2006.

\bibitem[CT06]{elements2006}
Thomas~M. Cover and Joy~A. Thomas.
\newblock {\em Elements of {I}nformation {T}heory}.
\newblock Wiley-Interscience, second edition, 2006.

\bibitem[DH00]{domingos2000mining}
Pedro Domingos and Geoff Hulten.
\newblock Mining {H}igh-{S}peed {D}ata {S}treams.
\newblock In {\em Proceedings of the {S}eventh {A}{C}{M} {S}{I}{G}{K}{D}{D}
  {I}nternational {C}onference on {K}nowledge {D}iscovery and {D}ata {M}ining},
  KDD '00, pages 71--80, New York, NY, USA, 2000. ACM.

\bibitem[HSD01]{hulten2001mining}
Geoff Hulten, Laurie Spencer, and Pedro Domingos.
\newblock Mining {T}ime-{C}hanging {D}ata {S}treams.
\newblock In {\em Proceedings of the {S}eventh {A}{C}{M} {S}{I}{G}{K}{D}{D}
  {I}nternational {C}onference on {K}nowledge {D}iscovery and {D}ata {M}ining},
  KDD '01, pages 97--106, New York, NY, USA, 2001. ACM.

\bibitem[Mit97]{mitchell1997ml}
Thomas~M. Mitchell.
\newblock {\em Machine {L}earning}.
\newblock McGraw-Hill, Inc., New York, NY, USA, 1st edition, 1997.

\bibitem[Sha48]{shannon48}
Claude~E. Shannon.
\newblock A {M}athematical {T}heory of {C}ommunication.
\newblock {\em The Bell System Technical Journal}, 27:379--423, 623--656, July,
  October 1948.

\bibitem[Sov13]{sovdat2013algorithms}
Bla\v{z} Sovdat.
\newblock {A}lgorithms for {I}ncremental {L}earning of {D}ecision {T}rees from
  {T}ime-{C}hanging {D}ata {S}treams.
\newblock Bachelor's thesis, University of Ljubljana, September 2013.

\bibitem[SS13]{MO133986}
Andr\'{e} Schlichting and Bla\v{z} Sovdat.
\newblock {I}ncremental entropy computation.
\newblock MathOverflow, 2013.
\newblock [Online; accessed 2013-06-17] Available at
  \url{http://mathoverflow.net/questions/133986}.

\end{thebibliography}

\end{document}